\documentclass[10pt,twocolumn,letterpaper]{article}

\usepackage[final]{iccv}      % To produce the REVIEW version
\usepackage{makecell}
\usepackage{stfloats}
\usepackage{multirow} 
\usepackage{colortbl}
\usepackage{xcolor}
\usepackage{graphicx}
\usepackage{caption}
\setcellgapes{0}
\definecolor{iccvblue}{rgb}{0.21,0.49,0.74}
\usepackage[pagebackref,breaklinks,colorlinks,allcolors=iccvblue]{hyperref}

\title{EgoSplat: Open-Vocabulary Egocentric Scene Understanding with Language Embedded 3D Gaussian Splatting}

\author{Di Li\\
Xidian University\\
{\tt\small dili@stu.xidian.edu.cn}
\and
Jie Feng\\
Xidian University\\
{\tt\small jiefeng0109@163.com}
\and
Jiahao Chen\\
Sun Yat-sen University\\
{\tt\small chenjh328@mail2.sysu.edu.cn}
\and
Weisheng Dong\\
Xidian University\\
{\tt\small wsdong@mail.xidian.edu.cn}
\and
Guanbin Li\\
Sun Yat-sen University\\
{\tt\small liguanbin@mail.sysu.edu.cn}
\and
Guangming Shi\\
Xidian University\\
{\tt\small gmshi@xidian.edu.cn}
\and
Licheng Jiao\\
Xidian University\\
{\tt\small lchjiao@mail.xidian.edu.cn}\\\
}

\begin{document}

\twocolumn[{
\maketitle
\begin{center}
    \captionsetup{type=figure}
    \includegraphics[width=1\textwidth]{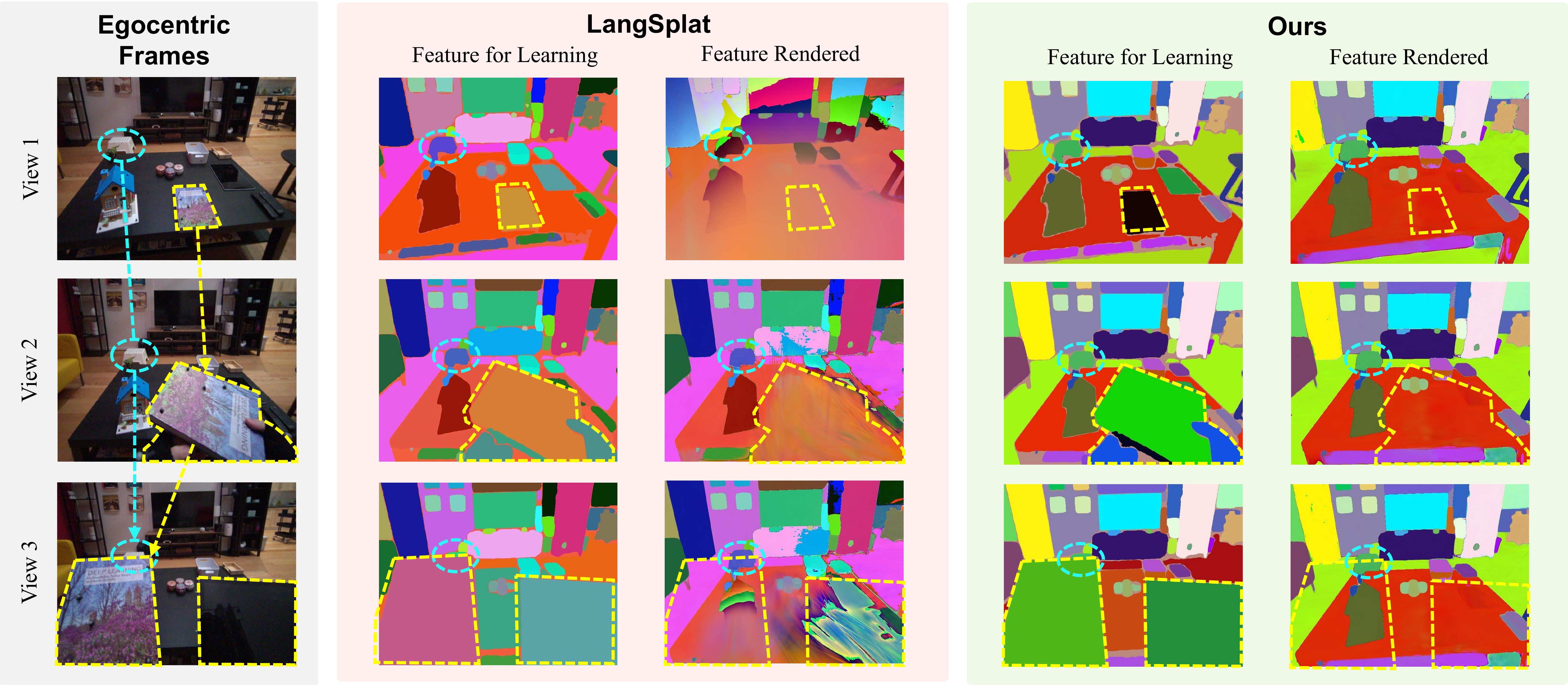}
       \caption{\textbf{Comparison of the visualization of language feature for 3DGS learning and the rendered feature between previous SOTA method Langsplat~\cite{qin2024langsplat} and the proposed Egosplat}.  With its multi-view instance feature aggregation and transient object modeling, the proposed EgoSplat achieves enhanced multi-view consistency (occlusions highlighted by \textcolor[rgb]{ 0,  1,  1}{cyan} circles) and minimized artifacts (transient objects highlighted by \textcolor[rgb]{ 1,  1,  0}{yellow} polygons) through transient filtering. 
   }
   \label{fig:teaser}
\end{center}
}]

\maketitle

\begin{abstract}
% Understanding the 3D visual world from human perspectives has been a long-standing challenge in computer vision. 
Developing a 3D language framework that enables open-vocabulary querying is crucial for  interacting with egocentric 3D scenes effectively.
%Developing a 3D language field that supports open-vocabulary queries is essential for egocentric 3D scenes. %
Egocentric scenes exhibit frequent occlusions, varied viewpoints, and dynamic interactions compared to typical scene understanding tasks. Occlusions and varied viewpoints can lead to multi-view semantic inconsistencies, while dynamic objects may act as transient distractors, introducing artifacts into semantic feature modeling.
To address these challenges, we propose EgoSplat, a language-embedded 3D Gaussian Splatting framework for open-vocabulary egocentric scene understanding. A multi-view consistent instance feature aggregation method is designed to leverage the segmentation and tracking capabilities of SAM2 to selectively aggregate complementary features across views for each instance, ensuring precise semantic representation of scenes. Additionally, an instance-aware spatial-temporal transient prediction module is constructed to improve spatial integrity and temporal continuity in predictions by incorporating spatial-temporal associations across multi-view instances, effectively reducing artifacts in the semantic reconstruction of egocentric scenes.
EgoSplat achieves state-of-the-art performance in both localization and segmentation tasks on two datasets, outperforming existing methods with a 8.2\% improvement in localization accuracy and a 3.7\% improvement in segmentation mIoU on the ADT dataset, and setting a new benchmark in open-vocabulary egocentric scene understanding. The code will be made publicly available.

\end{abstract}
    
\section{Introduction}
\label{sec:intro}

%-------------------------------------------------------------------------

% Egocentric 3D 

Egocentric 3D scene understanding focuses on perceiving, analyzing and interpreting the physical environment from a first-person perspective, making it essential for applications, such as augmented and virtual reality~\cite{grauman2022ego4d,gu2025egolifter,Liu_2022_CVPR}.
Recently, advanced 3D representation techniques, such as Neural Radiance Fields (NeRF)~\cite{mildenhall2020nerf} and 3D Gaussian Splatting (3DGS)~\cite{kerbl20233d}, have made it possible to efficiently reconstruct 3D scenes with spatial consistency for more accurate perception.
Additionally, there has been growing interest in enabling natural language-based interactions between users and their surroundings with the rise of large language models.

A promising direction is to combine the strengths of these techniques to model the semantic information of 3D egocentric scenes, facilitating open-vocabulary understanding and enhancing user interactions.
% However, combining these techniques is non-trivial because of the inherent challenges of egocentric scenes.
% Firstly, egocentric videos are constrained by a narrow field of view and limited camera movement, making it difficult to capture the diverse viewpoints needed for accurate scene reconstruction. 
% %
% Furthermore, frequent human-object interactions in egocentric videos result in dynamic occlusions, causing multi-view inconsistencies, and transient objects would introduce noise or artifacts into the modeling process.
This combination these techniques is non-trivial because of the inherent challenges of egocentric scenes.
Firstly, egocentric videos are constrained by a narrow field of view and limited camera movement, making it difficult to capture the diverse viewpoints needed for accurate semantic reconstruction. 
Furthermore, frequent human-object interactions in egocentric videos result in dynamic occlusions, causing multi-view semantic inconsistencies, and transient objects would introduce noise or artifacts into the modeling process~\cite{gu2025egolifter}.

%existing method
Recent works~\cite{kerr2023lerf,qin2024langsplat,shi2024language,peng2024gags} have investigated distilled vision features from pre-trained foundation models on non-egocentric scenes, like CLIP~\cite{radford2021learning}, OpenSeg~\cite{ghiasi2022scaling} and DINO~\cite{caron2021emerging}, by incorporating additional output branches in scene representation to predict language embedded features. Notably, LangSplat~\cite{qin2024langsplat} achieves state-of-the-art performance by integrating hierarchical semantics to obtain CLIP features.
However, these methods perform poorly in egocentric scenes due to their lack of consideration for frequent occlusions, varying viewpoints, and dynamic interactions. 

As illustrated in \cref{fig:teaser}, due to continuous interactions in the egocentric video, occluded objects acquire inconsistent language features, leading to semantic inconsistencies in the rendered feature, while improper handling of dynamic interactions introduces artifacts in the rendered feature of LangSplat.

%In this paper
In this paper, we propose the EgoSplat, which is the first language-embedded field method tailored for egocentric scenes.  
We start by a multi-view consistent instance feature aggregation method designed  by learning language features for each instance from CLIP, filtering out the unsatisfactory instance views that are prone to semantic conflict, and then aggregating the remaining high-quality instance views to extract complementary cross-view features for each instance.  In this process, the promptable visual foundation model, Segment Anything Model 2 (SAM2)~\cite{ravi2024sam}, is employed for for segmentation and  tracking on egocentric videos to generate the 
associated instance-level segmentation results.  
To handle dynamic objects, an instance-aware spatial-temporal transient prediction module is constructed to enhance the transient  spatial integrity and temporal continuity  through a spatial-temporal instance-aligened way. Subsequently, the dynamic elements are filtered out  to mitigate the interference from scene dynamics. 
Experiments on the Aria Digital Twin (ADT) dataset~\cite{pan2023aria} and the HOI4D dataset~\cite{Liu_2022_CVPR} demonstrate the superiority of our method.
% %contributions
Our contributions can be summarized as follows:

\begin{itemize}
    \item EgoSplat introduces a paradigm shift by transitioning from viewpoint-specific features for semantic reconstruction to aggregating cross-viewpoint features across temporal and spatial dimensions, enabling open-vocabulary semantic understanding of highly dynamic, occluded, and varied egocentric scenes.
    % , by leveraging instance-wise spatial and temporal associations from a 3D modeling perspective. 
    % \item EgoSplat introduces a paradigm shift by moving from viewpoint-independent features for semantic reconstruction to aggregating viewpoint-specific features across temporal and spatial dimensions.Enabling open-vocabulary semantic understanding of highly dynamic, occluded, and varied scenes through instance-wise spatial and temporal associations from a 3D modeling perspective.

%We propose the EgoSplat, which is the first field-based method for open-vocabulary  scene understanding of natural dynamic egocentric videos. 
    \item The multi-view consistent instance feature aggregation method eliminates semantic conflict caused by occlusions through high-quality view selection, and improves the semantic feature representation ability through cross-view consistency semantic aggregation.
%A multi-view consistent instance feature aggregation method is designed to filter out anomalous viewpoints and aggregate remaining view features to obtain precise language features.
%An instance-aware spatial-temporal transient prediction module is proposed to filter out dynamic components, thereby providing robust mitigating transient distractors.
    \item  The instance-aware spatial-temporal transient prediction module enhances the boundary integrity of dynamic objects, and improves the accuracy of transient predictions, which  suppresses the interference of dynamic elements and enhances the quality of reconstruction effectively. 
    % \item Experimental results on two natural dynamic egocentric datasets demonstrate that our method surpasses state-of-the-art approaches significantly in novel-view 3D open-vocabulary localization and segmentation tasks. 其中 \textbf{relative performance} gains of 23.8\% in localization and 46.2\% in segmentation, demonstrating substantial progress in egocentric scene understanding.
    \item Experimental results on two dynamic egocentric datasets demonstrate that our method outperforms state-of-the-art approaches, achieving absolute performance improvements of 8.2\% and 3.7\%, as well as relative performance gains of 29.6\% and 35.6\% in novel-view 3D open-vocabulary localization and segmentation tasks, respectively.
\end{itemize}

%contributions
\section{Related Work}
\label{sec:relate}

%-------------------------------------------------------------------------
\begin{figure*}[t]
  \centering
  \includegraphics[width=1\textwidth]{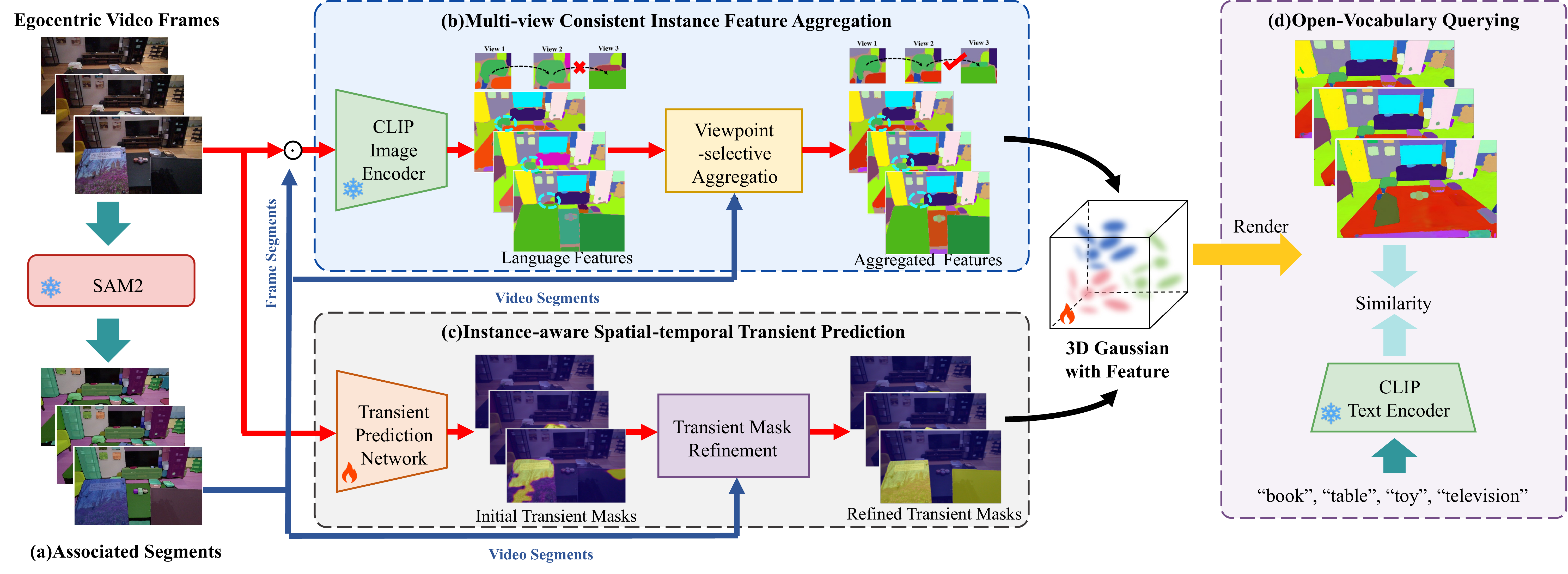}

\caption{\textbf{Pipeline of EgoSplat.} Given a sequence of egocentric video frames, (a) SAM2 performs video segmentation to obtain associated segments for each instance across frames. (b) A multi-view consistent instance feature aggregation module that selects high-quality views is employed to extract precisely  language features. (c) For dynamic objects, an instance-aware spatial-temporal transient prediction module is designed to achieve transient prediction with temporal continuity and spatial completeness. We then train 3D Gaussians using consistent instance features with dynamic objects filtered out. (d) During querying, the similarity between rendered language embeddings and text embeddings enables open-vocabulary localization and segmentation through natural language interaction.}
  \label{fig:pipline}
\end{figure*}

\vspace{1mm}
\noindent \textbf{Open-Vocabulary Scene Understanding.}
% Egocentric scene understanding, aimed at developing AI agents capable of perceiving and interpreting the surrounding environment, has long been a challenge in computer vision.
Egocentric scene understanding has long been a challenge in computer vision.
Early approaches focused on identifying specific elements through category specification~\cite{zhang2022fine,tokmakov2023breaking,cheng2023towards,zhu2023egoobjects}, visual querying~\cite{xu2023my,mai2023egoloc,jiang2024single} and self-supervised learning~\cite{akiva2023self}.
With the help of vision-language models, recent works~\cite{nagarajan2023egoenv,chatterjee2024opening,jia2022egotaskqa,lin2022egocentric,pramanick2023egovlpv2} are able to perform scene understanding based on natural language prompts, providing users with more convenient interactions. This progression introduces the concept of open-vocabulary scene understanding, where models are not confined to a predefined set of categories but can recognize and interpret a wide array of objects and scenes based on natural language descriptions. 
However, most of these methods are designed for 2D images or video sequences and often suffer from spatial inconsistencies due to the lack of 3D modeling, which is crucial for accurate scene representation and understanding.

\vspace{1mm}
\noindent \textbf{3D Representations in Egocentric Scenes.}
% As one of the latest 3D representations, 3D Gaussians Splatting (3DGS)~\cite{kerbl20233d} conceptualizes a static scene as an explicit Gaussian point cloud, providing enhanced visual quality and improved rendering speed over Neural Radiance Fields (NeRF)~\cite{mildenhall2020nerf}.
% Since the real world is non-static, several recent works have been proposed to extend 3DGS to reconstruct entire dynamic scenes.
% Most of these methods~\cite{luiten2024dynamic,wu20244d,huang2024s3,li2024spacetime,yang2024deformable} attempt to use neural fields or motion functions to represent the deformation of 3D Gaussians at different timesteps, while others~\cite{duan20244d,yang2023real} extend 3D Gaussians to 4D with additional time-based dimensions.
% Other works treat real-world scenes as static, with dynamic objects decoupled to be modeled separately~\cite{yan2024street,chen2024omnire,wu2022d,yang2023emernerf,zhou2024drivinggaussian,zhou2024hugs} or eliminated as transient distractors~\cite{martin2021nerf,sabour2023robustnerf,sabour2024spotlesssplats,chen2024nerf,ren2024nerf,kulhanek2024wildgaussians,zhang2024gaussian,chen2022hallucinated}.
As a recent 3D representation, 3D Gaussians Splatting (3DGS)~\cite{kerbl20233d} models a static scene as Gaussian point clouds, providing better performance over Neural Radiance Fields (NeRF)~\cite{mildenhall2020nerf}.
To extend 3DGS for dynamic scene reconstruction, several methods~\cite{luiten2024dynamic,wu20244d,huang2024s3,li2024spacetime,yang2024deformable} introduce motion functions to represent the deformation of 3D Gaussians, while others~\cite{duan20244d,yang2023real} extend 3D Gaussians to 4D with additional time-based dimensions.
Additionally, some approaches treat dynamic objects separately from the static scene, either modeling them independently~\cite{yan2024street,chen2024omnire,wu2022d,yang2023emernerf,zhou2024drivinggaussian,zhou2024hugs} or eliminating them as transient distractors~\cite{martin2021nerf,sabour2023robustnerf,sabour2024spotlesssplats,chen2024nerf,ren2024nerf,kulhanek2024wildgaussians,zhang2024gaussian,chen2022hallucinated}.

Nonetheless, the approaches mentioned-above typically rely on multi-view images from multiple static cameras, making them unsuitable for egocentric scenes~\cite{sun2023aria,tschernezki2024epic,pan2023aria,Liu_2022_CVPR,damen2018scaling}.
Egocentric scenes are captured by a limited number of cameras with a narrow baseline and involve complex interactions between actors and objects.
To address this challenge, NeuralDiff~\cite{tschernezki2021neuraldiff} and N3F~\cite{tschernezki2022neural} use separate NeRF branches to implicitly distinguish different dynamic elements from the static background and reconstruct them individually.
EgoGaussian~\cite{zhang2024egogaussian} employs existing segmentation algorithms to explicitly identify hand-object interactions and estimates object motions.
In contrast, EgoLifter~\cite{gu2025egolifter} focuses on static environments and intentionally excludes dynamic elements by applying a 2D prediction network.
In comparison to these methods, our method integrates the state-of-the-art promptable video segmentation model~\cite{ravi2024sam} guided by the transient prediction network to leverage the instance-aware spatial-temporal information of egocentric videos, thereby achieving accurate tracking and segmentation of transient elements across frames.

\vspace{1mm}
\noindent \textbf{Language Embedded 3D Representations.}
3D representations with semantic information have gained significant attention for 3D scene understanding.
Several works~\cite{zhi2021place,kobayashi2022decomposing,siddiqui2023panoptic,liu2023weakly,kerr2023lerf,engelmann2024opennerf,yang2023emernerf,fan2022nerf,goel2023interactive} have explored distilling 2D vision features from pre-trained foundation models~\cite{caron2021emerging,radford2021learning,oquab2023dinov2,kirillov2023segment,li2022language,ghiasi2022scaling} into NeRF frameworks.
Notably, LERF~\cite{kerr2023lerf}, 3D-OVS~\cite{liu2023weakly} and OpenNeRF~\cite{engelmann2024opennerf} enable 3D open-vocabulary querying by embedding features from multimodal models (\eg, CLIP~\cite{radford2021learning}, LSeg~\cite{li2022language} and OpenSeg~\cite{ghiasi2022scaling}).
Although attempts have been made to replace NeRF with 3DGS for better results in static~\cite{qin2024langsplat,zhou2024feature,zuo2024fmgs,wu2024opengaussian,ye2025gaussian} and dynamic~\cite{labe2024dgd,guo2024semantic} scenes, they have not been validated in egocentric settings and demonstrate limited performance in our experiments.
Other methods~\cite{tschernezki2021neuraldiff,tschernezki2022neural,gu2025egolifter} designed for egocentric scenes cannot achieve open-vocabulary understanding due to the lack of language features.
To our knowledge, our work is the first to apply language embedded 3DGS to egocentric scenes and improve its performance by leveraging video temporal consistency.
\section{Proposed Approach}
\label{sec:method}

To address the challenges of open-vocabulary scene understanding in egocentric scenarios, we integrate the capabilities of 3DGS with vision-language models, introducing EgoSplat, a novel language-embedded 3DGS framework. The pipeline of our proposed method is illustrated in \cref{fig:pipline}. Note that, to enhance readability, we have visualized frames with smaller viewpoint variations.
To overcome issues such as frequent occlusions and varied viewpoints, we propose a multi-view consistent instance feature aggregation method. To  accurately and completely predict dynamic targets, we introduce an instance-aware spatio-temporal transient prediction module. The 3D Gaussian features are then trained and subsequently rendered, enabling open-vocabulary localization and segmentation.

\subsection{Preliminary: 3DGS with Language Features}
3DGS~\cite{kerbl20233d} represents scenes using the collections of 3D Gaussians, where each Gaussian $G_i$ is characterized  by its position $\mu_i$, covariance $\Sigma_i$, color $c_i$, and opacity $\alpha_i$. The rendering involves projecting these Gaussians onto the image space, facilitating efficient real-time synthesis of novel views. To incorporate semantic information, we associate the semantic embedding $f_i$ with each Gaussian. Thus, each Gaussian $G_i$ is represented as:
\begin{equation}
    \label{equ:gaussian} G_i = \{ \mu_i, \Sigma_i, c_i, \alpha_i, f_i \}.
\end{equation}
Rendering in 3DGS entails projecting all 3D Gaussians onto the 2D image plane according to camera
parameters, producing the rendered image $\hat{I}$ and the language feature map $\hat{F}$. Training 3DGS optimizes the Gaussian parameters to minimize discrepancies between rendered and ground-truth images and language feature maps across multiple viewpoints. The photometric reconstruction loss $\mathcal{L}_{\text{RGB}}$ is defined as follows:
\begin{equation}
    \label{equ:rgb_loss}
   \mathcal{L}_{\text{RGB}} = \sum_{p \in \mathcal{P}}  d_{I}( I(p) - \hat{I}(p)) ,
\end{equation}
where $\mathcal{P}$ denotes pixel locations of ground-truth image $I$ and rendered image $\hat{I
}$, $d_{I}$ denotes the distance function used for the image. An additional semantic consistency loss $L_{\text{F}}$ aligns the embedding $f_i$ with corresponding visual features:
\begin{equation}
    \label{equ:fea_loss}
\mathcal{L}_{\text{F}} = \sum_{p \in \mathcal{P}} d_{F}( F(p) - \hat{F}(p) ),
\end{equation}
where \( d_{F}\) denotes the distance function used of features for 3DGS with language features, \(F\) and \(\hat{F}\) denotes the  ground truth and rendered language features.
The total loss function is:
\begin{equation}
    \label{equ:total_loss}
    \mathcal{L} = \mathcal{L}_{\text{RGB}} + \lambda_1\mathcal{L}_{\text{F}},
\end{equation}
where $\lambda_1$ balances the contributions of both losses. Minimizing $\mathcal{L}$ with optimization techniques, like stochastic gradient descent, allows the model to effectively capture both visual appearance and semantic content. 
This integration of 3DGS with language features enables intuitive open-vocabulary querying and accurate semantic segmentation within 3D scenes, enhancing both visual and semantic understanding for a range of interactive applications.

\subsection{Multi-View Consistent Instance Feature Aggregation}
We address the challenges of semantic inconsistency caused by frequent occlusions and viewpoint variations by performing instance segmentation and association across viewpoints, removing anomalous perspectives, and then aggregating features from the remaining viewpoints. To achieve this, SAM2 is first applied , which offers flexibility in obtaining segmentations that capture cross-viewpoint relationships in video. After that, we propose a multi-view consistent instance feature aggregation method that leverages temporal instance information from SAM2 to selectively aggregate instance-level features across multiple viewpoints, effectively excluding views that could introduce semantic conflicts.

\vspace{1mm}
\noindent \textbf{Video Segmentation.} 
We first employ SAM2 to perform instance segmentation on videos. This process yields instance masks for all the objects in each frame, along with associated instance segments that track these objects across frames. Specifically, we begin by applying grid of point prompts to the SAM2 on the initial frame to generate object masks. Subsequently, we  remove redundant masks for each frame based on the predicted IoU score, stability score, and overlap rate between masks. Remaining  masks are then utilized as prompts, propagating through the entire video sequence to obtain corresponding segments $\mathcal{S}=\{S_i^j\}$ across all frames, where $S_{i}^{j} \in \mathbb{R}^{H\times W}$ represents the segment mask of the $i$-th instance in the $j$-th frame. 

\vspace{1mm} 
\noindent \textbf{View-Consistent Feature extraction.} 
With the segmented instances from the video frames, we proceed to extract CLIP features for each instance across all frames. The viewpoint-specific language feature $F_{i}^{j} \in\mathbb{R}^{\text{ndim}}$ of the $i$-th instance with \(ndim\) dimension in the $j$-th frame can be formulated as:
\begin{equation}
    \label{equ:clip_feature}
     F_{i}^{j} = \text{CLIP}(S_{i}^{j} \odot I_{i}^{j}),
\end{equation}
where \(\odot \) denotes the mask and crop operation. 
% following LangSplat~\cite{qin2024langsplat}.

To filter out viewpoints with features caused by occlusions or perspective changes, we  employ the Median Absolute Deviation (MAD)~\cite{law1986robust} anomaly detection algorithm to identify and exclude semantically conflicting features. Specifically, for each instance, we calculate its median feature vector $\tilde{F}_i$ across all frames:
\begin{equation}
    \label{equ:mad_fi}
\tilde{F}_i = \text{median}\left( \{ F_i^j \}_{j=1}^N \right),
\end{equation}
where $N$ is the number of frames. The MAD for the $i$-th instance is then computed as:
\begin{equation}
    \label{equ:mad}
\text{MAD}_i = \text{median}\left( \left\{ \left| F_i^j - \tilde{F}_i \right| \right\}_{j=1}^N \right).
\end{equation}

The view-consistent feature \( \bar{F}_{i} \) for instance \( i \) is computed as the average of the non-conflict features:
\begin{equation}
    \label{equ:aggr_feature}
\bar{F}_{i} = \frac{1}{N'} \sum_{j=1}^{N} F_{i}^{j} \cdot \mathbb{I}\left( \left| F_i^j - \tilde{F}_i \right| \leq k \times \text{MAD}_i \right),
\end{equation}
where \( \mathbb{I}(\cdot) \) is the indicator function, \( N' \) is the number of non-anomalous features,  and the feature vector \( F_i^j \) is considered anomalous if it exceeds a \( k \) times MAD threshold. This approach ensures that \( \hat{F}_{i} \) is robust against anomalies caused by occlusions or incomplete views, thereby enhancing the consistency and reliability of the extracted features. To retain certain view-specific semantic information, the final feature acted as ground truth is obtained by averaging the view-specific feature $F_i^j$ and the view-independent feature $\bar{F}_{i}$. 

% \subsection{Spatial-Temporal-Aware Transient Prediction}
\subsection{Instance-Aware Spatial-Temporal Transient Prediction}

%Segmentation-Refined Transient Prediction
% As mentioned above in \cref{sec:relate}, 
Dynamic actor-object interactions are another challenge in egocentric videos that must be properly handled, otherwise they will act as transient distractors and introduce artifacts into the rendered results.
To achieve egocentric scene understanding, we focus on the static objects by ignoring transient elements instead of reconstructing them. Thus, accurate prediction of dynamic objects is crucial in the egocentric scenes.  Specifically, let $M\in \mathbb{R}^{H \times W}$ be the probability map corresponding to image $I$, which assigns the probability of each pixel belonging to transient elements. We modify \cref{equ:rgb_loss} and \cref{equ:fea_loss}  by using \(1-M\) as the loss weight to reduce the interference of pixels with higher transient probabilities:
\begin{equation}
    \label{equ:rgb_w_loss}
    \mathcal{L}_{\text{RGB-w}}(I, \hat{I}, M) = \sum_{p \in \mathcal{P}} \left(1-M(p)\right) d_{I}( I(p) - \hat{I}(p))
\end{equation}
\begin{equation}
    \label{equ:feat_w_loss}
    \mathcal{L}_{\text{F-w}}(F, \hat{F}, M) = \sum_{p \in \mathcal{P}} \left(1-M(p)\right) d_{F}( F(p) - \hat{F}(p))
\end{equation}
\begin{figure}[t]
  \centering
   \includegraphics[width=1\linewidth]{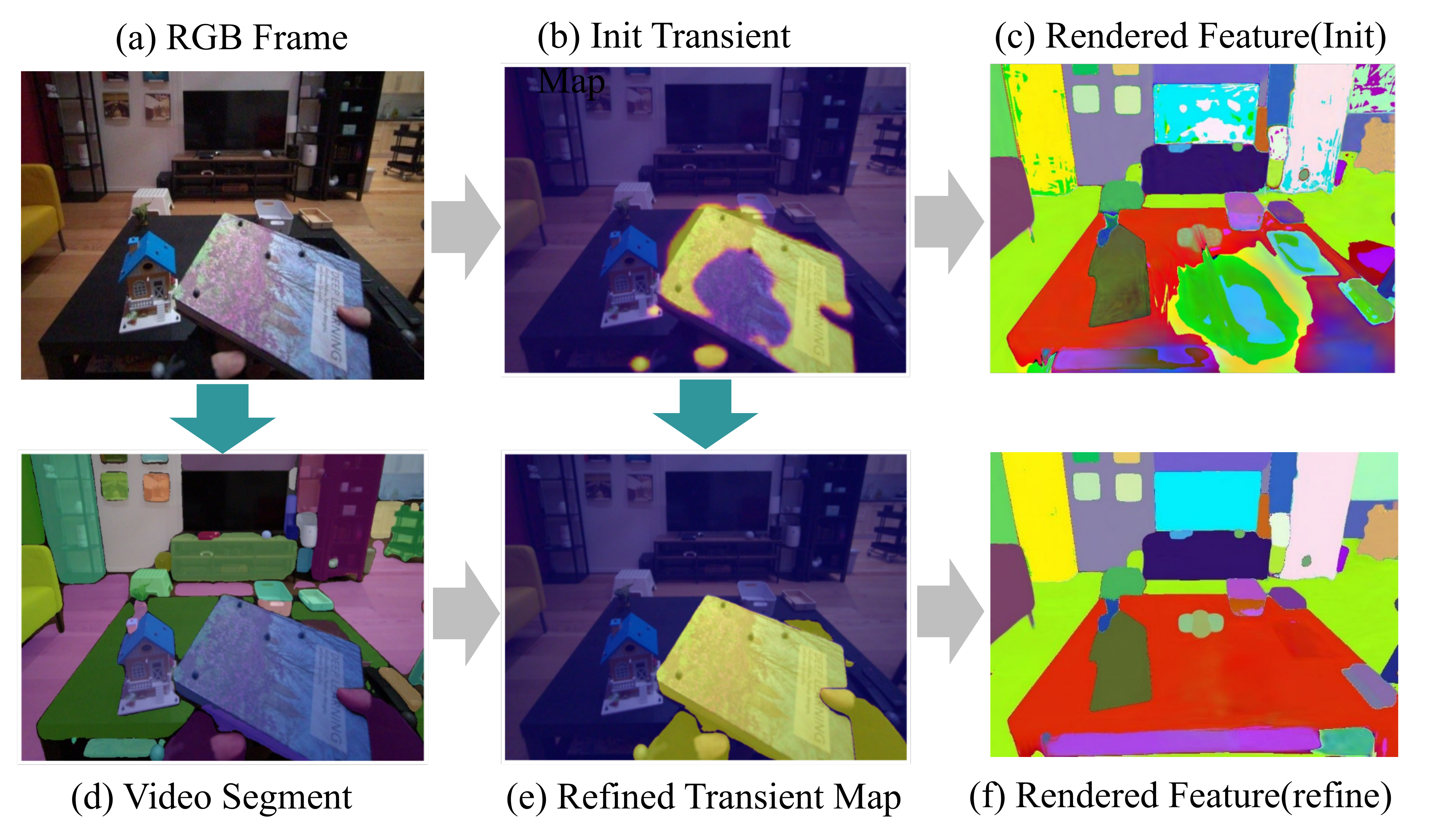}

   \caption{\textbf{Instance-aware spatial-temporal transient prediction}. (a) RGB Frame. (b) Initial transient map generated by a 2D transient prediction network. (c) Rendered features using the initial transient map, where incomplete masks introduce artifacts. (d) Video segments from SAM2. (e) Refined transient map using video segments, resulting in masks with improved edge definition. (f) Rendered feature using the refined transient map.
   }
   \label{fig:trans_pred}
\end{figure}
% To obtain $M$, directly employ a learnable prediction network to output the probability maps used in the 3DGS training process. Despite their effectiveness, these methods often produce imprecise or erroneous results, mainly due to their incomplete utilization of spatio-temporal information.
\noindent To obtain $M$, directly employing a learnable prediction network to output probability maps for the 3DGS training process is somewhat effective~\cite{gu2025egolifter}. However, it often produces imprecise or erroneous results, primarily due to the incomplete utilization of spatio-temporal information.
We instead  propose a two-stage transient prediction approach: 1) The prediction network is first optimized to generate an initial transient map in the \textit{initialization stage}, and then 2) the initial transient map is refined using the promptable video segmentation result  to produce more accurate map in the subsequent \textit{spatial-temporal refinement stage}.

\vspace{1mm}
\noindent \textbf{Initialization Stage.}
We adopt a learnable 2D transient prediction network $H$, which takes the training image $I$ as input and outputs the corresponding transient probability map $M$.
To optimize it, the loss function $\mathcal{L}_{\text{init}}$ is used with the assistance of the rendered image $\hat{I}$ from a temporary 3DGS:
\begin{equation}
    \mathcal{L}_{\text{init}} = \mathcal{L}_{\text{RGB-w}}(I, \hat{I}, M) + \lambda_2 \sum_{p \in \mathcal{P}} \left| M(p)\right|,
    \label{epu:trans_l_init}
\end{equation}
where the second term scaled by $\lambda_2$ is a $L_1$ regularizer following the existing works~\cite{gu2025egolifter,martin2021nerf,zhang2024gaussian}.
After optimization, the network $H$ is freezed and can roughly identify transient elements in each training image.

\vspace{1mm}
\noindent \textbf{Spatial-Temporal Refinement Stage.}
Instead of directly using the predicted map $M^j=H(I^j)$ for the $j$-th training frame $I^j$, we treats it as prompts for SAM2 to accurately segment transient  instances  across all frames.
Specifically, we set a threshold $\mathcal{T}$ to obtain the set $\mathcal{D}$ of transient element indices:
\begin{equation}
    \mathcal{D} = \left\{i \left\vert\ \frac{\sum_{j=1}^{N_I} S_i^jM^j}{\sum_{j=1}^{N_I}S_i^j} \geq \mathcal{T}\right.\right\},
\end{equation}
and then we have the sharp and refined transient mask $\hat{M}^j$ for $I^j$, incorporating the instance-level spatio-temporal information introduced by SAM2: 
\begin{equation}
    \hat{M}^j = \bigcup_{i\in\mathcal{D}} S_i^j.
\end{equation}

Finally, we apply these refined transient masks and the loss function modified from \cref{equ:total_loss} to train EgoSplat:
\begin{equation}
    \mathcal{L} = \mathcal{L}_{\text{RGB-w}}(I, \hat{I}, \hat{M}) + \lambda_1\mathcal{L}_{\text{F-w}}(F, \hat{F}, \hat{M}). 
\end{equation}
Therefore, the spatio-temporal information from SAM2 instances is utilized to refine the transient prediction.

% \subsection{Open-Vocabulary Localization and Segmentation}
\subsection{Open-Vocabulary Querying}
Since EgoSplat is embedded with language features from CLIP that has aligned the latent space between images and texts, it can render view-specific feature maps to achieve 3D open-vocabulary querying, such as object localization and semantic segmentation.

\vspace{1mm}
\noindent \textbf{Object Localization.}
To localize the objects described by text queries in the egocentric scene, we follow LERF~\cite{kerr2023lerf} and LangSplat~\cite{qin2024langsplat} to 
compute the relevancy score $R$ between the specific rendered pixel feature $F$ and the encoded text query feature $F_{\text{q}}$:
\begin{equation}
    R = \min_{F_{\text{c}}\in \mathcal{F}_\text{c}}\frac{\exp\left(\mathop{sim}\left(F, F_{\text{q}}\right)\right)}{\exp\left(\mathop{sim}\left(F, F_{\text{q}}\right)\right) + \exp\left(\mathop{sim}\left(F, F_{\text{c}}\right)\right)},
\end{equation}
where $\mathop{sim}(\cdot,\cdot)$ is the cosine similarity, and $\mathcal{F}_{\text{c}}$ is the encoded feature set of predefined canonical phrases (\ie, ``\textit{object}'', ``\textit{things}'', ``\textit{stuff}'' and ``\textit{texture}'').
The point with the highest relevancy score in the entire rendered image will be selected as the localization of the target object.

\vspace{1mm}
\noindent \textbf{Semantic Segmentation.}
For the 3D semantic segmentation task, we flow LERF~\cite{kerr2023lerf} and LangSplat~\cite{qin2024langsplat} to  filter out points with relevancy scores lower than a chosen threshold, and predict the object masks with remaining regions.

\section{Experiments}
\label{sec:exp}

% \vspace{3pt}
\begin{figure}[t]
  \centering
   \includegraphics[width=1\linewidth]{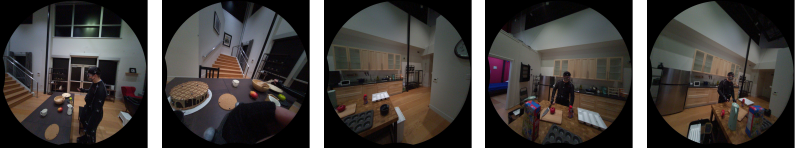}
    % \vspace{-22pt} % 减小图像和 caption 之间的距离
   \caption{Typical frames from Scene 2 in the ADT dataset. Egocentric scenes are characterized by a narrow field of view, various camera coverage, and frequent human-object interactions. }
   \label{fig:adt_sample}
   % \vspace{-3mm}
\end{figure}

\subsection{Settings}

\noindent \textbf{Implementation Details.} For video segmentation, we employ the \texttt{hiera\_base\_plus} SAM2 model, utilizing a \(32 \times 32\) point prompt for each image. 

For language feature extraction, we follow the methodology outlined in LangSplat~\cite{qin2024langsplat}, employing the OpenCLIP ViT-B/16 model to obtain CLIP features at three distinct SAM2 scales. In the multi-view consistent instance feature aggregation process, the threshold \( k \) in the MAD algorithm is set to 3. To prevent GPU out-of-memory errors due to the high dimensionality of CLIP features, we utilize the autoencoder from LangSplat~\cite{qin2024langsplat} to compress the features to 3 dimensions, adhering to the same settings.
Regarding the instance-aware spatial-temporal transient prediction, we use the same transient prediction network as EgoLifter's~\cite{gu2025egolifter}.
We train the prediction network for 7,000 iterations in the initialization stage, followed by training EgoSplat for 30,000 iterations in the refinement stage.
All experiments are conducted on a single NVIDIA RTX-3090 GPU. Additional details are provided in the supplementary materials.

% \vspace{1mm} 
\noindent \textbf{Datasets.} We employ two datasets for evaluation:
\begin{itemize} \item \textit{Aria Digital Twin (ADT)}~\cite{pan2023aria} is an egocentric dataset collected using Aria glasses, offering extensive ground truth annotations for devices, objects, and environments. It comprises 200 sequences of real-world activities conducted by Aria users in two indoor settings—a furnished apartment and an office space—featuring 398 object instances (344 stationary and 74 dynamic). ADT provides detailed ground truth data, including 3D bounding boxes and 2D segmentation masks for all frames. As ADT does not have a predefined setup for open-vocabulary localization or segmentation, we randomly selected 5 videos in ADT and reprocessed the original annotations to create the evaluation benchmark, generating GT 2D masks suitable for our tasks. % We will release the evaluation benchmark. Further details are available in the supplementary materials.
% The evaluation set of each scene in ADT was constructed through equidistant sampling, ensuring that the distribution of viewpoint variations is consistent with that of the entire dataset, as shown in \cref{fig:reb_1}.
Typical frames from a scene in the ADT dataset is shown in \cref{fig:adt_sample}.

\item \textit{HOI4D}~\cite{Liu_2022_CVPR} is an egocentric dataset emphasizing human-object interactions, with comprehensive frame-by-frame annotations for tasks such as action segmentation, motion segmentation, panoptic segmentation, 3D hand pose estimation, and category-level object pose tracking. It includes over 4,000 sequences captured by 9 participants interacting with 800 distinct object instances across 16 categories in 610 different indoor environments. The original 2D panoptic segmentation annotations in HOI4D were considered too coarse for open-vocabulary localization or segmentation tasks. Therefore, we re-annotated 5 scenes in this dataset to develop an evaluation benchmark tailored to these tasks. % The evaluation benchmark will be released. Additional information is provided in the supplementary materials.

\end{itemize}
The evaluation benchmark will be released, and further details are available in the supplementary materials.

\begin{table*}[htbp]
  \centering
  %   \centering
  \footnotesize
  \setlength{\tabcolsep}{3pt} % 设置列间距
  \caption{\textbf{Localization Accuracy(\%) on ADT dataset.}  The \textcolor[rgb]{ 1,  0,  0}{\textbf{1st}} and \textcolor[rgb]{ 0,  .502,  .502}{\textbf{2nd}} best results are highlighted.  ``St." and ``Dy." correspond to static and dynamic targets, respectively.}
    \begin{tabular}{l|ccc|ccc|ccc|ccc|ccc|ccc}
    \toprule
    \multicolumn{1}{c|}{\multirow{2}[4]{*}{Test scene}} & \multicolumn{3}{c|}{scene 1} & \multicolumn{3}{c|}{scene 2} & \multicolumn{3}{c|}{scene 3} & \multicolumn{3}{c|}{scene 4} & \multicolumn{3}{c|}{scene 5} & \multicolumn{3}{c}{avg} \\
\cmidrule{2-19}          & St.   & Dy.   & All   & St.   & Dy.   & All   & St.   & Dy.   & All   & St.   & Dy.   & All   & St.   & Dy.   & All   & St.   & Dy.   & All \\
    \midrule
    LERF~\cite{li2022language}  & 29.7  & 13.0  & 12.6  & 15.7  & 12.2  & 14.7  & 22.4  & 22.1  & 21.7  & 17.6  & 9.9   & 13.9  & 17.4  & 0.0   & 10.4  & 20.6  & 11.4  & 14.7  \\
    LEGaussian~\cite{shi2024language} & 18.0  & 10.9  & 16.5  & 18.0  & 3.7   & 11.7  & 18.8  & 14.7  & 18.2  & 10.4  & 11.4  & 11.8  & 17.4  & 17.2  & 18.1  & 16.5  & 11.6  & 15.3  \\
    SG~\cite{guo2024semantic}     & 23.4  & 10.9  & 19.7  & \textcolor[rgb]{ 0,  .502,  .502}{\textbf{19.1 }} & 13.4  & 17.1  & 21.2  & 19.1  & 21.7  & 28.0  & 9.1   & 22.0  & \textcolor[rgb]{ 1,  0,  0}{\textbf{25.9 }} & 5.7   & \textcolor[rgb]{ 0,  .502,  .502}{\textbf{21.9 }} & 25.1  & 11.6  & 20.5  \\
    GAGS~\cite{peng2024gags}  & 27.9  & 26.1  & 30.2  & 15.7  & \textcolor[rgb]{ 0,  .502,  .502}{\textbf{25.6 }} & 16.6  & \textcolor[rgb]{ 0,  .502,  .502}{\textbf{30.9 }} & 29.4  & \textcolor[rgb]{ 0,  .502,  .502}{\textbf{33.6 }} & 29.6  & \textcolor[rgb]{ 0,  .502,  .502}{\textbf{30.7}}  &  \textcolor[rgb]{ 0,  .502,  .502}{\textbf{32.8}}  & 17.4  & \textcolor[rgb]{ 0,  .502,  .502}{\textbf{25.3 }} & 21.8  & 24.3  & \textcolor[rgb]{ 0,  .502,  .502}{\textbf{27.4 }} & 27.0  \\
    Langsplat~\cite{qin2024langsplat} & \textcolor[rgb]{ 0,  .502,  .502}{\textbf{32.9 }} & \textcolor[rgb]{ 0,  .502,  .502}{\textbf{26.1 }} & \textcolor[rgb]{ 0,  .502,  .502}{\textbf{32.7 }} & 16.9  & 19.5  & \textcolor[rgb]{ 0,  .502,  .502}{\textbf{19.0 }} & 30.6  & \textcolor[rgb]{ 0,  .502,  .502}{\textbf{32.4 }} & 32.9  & \textcolor[rgb]{ 0,  .502,  .502}{\textbf{32.2 }} & \textcolor[rgb]{ 1,  0,  0}{\textbf{32.3 }} & 32.3  & 21.8  & 21.8  & 21.8  & \textcolor[rgb]{ 0,  .502,  .502}{\textbf{26.9 }} & 26.4  & \textcolor[rgb]{ 0,  .502,  .502}{\textbf{27.7 }} \\
    Ours  & \textcolor[rgb]{ 1,  .0,  .0}{\textbf{34.2 }} & \textcolor[rgb]{ 1,  0,  0}{\textbf{28.3 }} & \textcolor[rgb]{ 1,  .0,  .0}{\textbf{35.2 }} & \textcolor[rgb]{ 1,  0,  0}{\textbf{30.3 }} & \textcolor[rgb]{ 1,  0,  0}{\textbf{35.4 }} & \textcolor[rgb]{ 1,  0,  0}{\textbf{34.4 }} & \textcolor[rgb]{ 1,  .0,  .0}{\textbf{31.8 }} & \textcolor[rgb]{ 1,  0,  0}{\textbf{39.7 }} & \textcolor[rgb]{ 1,  0,  0}{\textbf{37.8 }} & \textcolor[rgb]{ 1,  0,  0}{\textbf{41.6 }} & 22.7  & \textcolor[rgb]{ 1,  0,  0}{\textbf{36.9 }} & \textcolor[rgb]{ 0,  .502,  .502}{\textbf{25.2 }} & \textcolor[rgb]{ 1,  0,  0}{\textbf{44.8 }} & \textcolor[rgb]{ 1,  0,  0}{\textbf{35.2 }} & \textcolor[rgb]{ 1,  0,  0}{\textbf{32.6 }} & \textcolor[rgb]{ 1,  0,  0}{\textbf{34.2 }} & \textcolor[rgb]{ 1,  0,  0}{\textbf{35.9 }} \\
    \bottomrule
    \end{tabular}%
  \label{tab:adt_seg}%
\end{table*}%

\begin{table*}[htbp]
  \centering
  \footnotesize
  \setlength{\tabcolsep}{3pt} % 设置列间距
  
\caption{\textbf{Segmentation IoU(\%) on the ADT dataset.}  The \textcolor[rgb]{ 1,  0,  0}{\textbf{1st}} and \textcolor[rgb]{ 0,  .502,  .502}{\textbf{2nd}} best results are highlighted.  ``St." and ``Dy." correspond to static and dynamic targets, respectively.}
    \begin{tabular}{l|ccc|ccc|ccc|ccc|ccc|ccc}
    \toprule
    \multicolumn{1}{c|}{\multirow{2}[3]{*}{Test scene}} & \multicolumn{3}{c|}{scene 1} & \multicolumn{3}{c|}{scene 2} & \multicolumn{3}{c|}{scene 3} & \multicolumn{3}{c|}{scene 4} & \multicolumn{3}{c|}{scene 5} & \multicolumn{3}{c}{avg} \\
\cmidrule{2-19}          & St.   & Dy.   & All   & St.   & Dy.   & All   & St.   & Dy.   & All   & St.   & Dy.   & All   & St.   & Dy.   & All   & St.   & Dy.   & All \\
\midrule
    LERF\cite{li2022language}   & 5.5   & 4.3   & 4.9   & 6.0   & 4.9   & 5.7   & 5.6   & 6.0   & 6.1   & 5.4   & 1.6   & 4.4   & 4.7   & 1.5   & 3.5   & 5.4   & 3.7   & 4.9  \\
    LEGaussian\cite{shi2024language} & 4.6   & 3.4   & 4.4   & 4.1   & 4.3   & 4.4   & 4.2   & 3.9   & 4.3   & 3.7   & 2.1   & 3.3   & 3.9   & 5.2   & 4.7   & 4.1   & 3.8   & 4.2  \\
    SG\cite{guo2024semantic}    & 7.2   & 5.4   & 7.0   & \textcolor[rgb]{ 0,  .502,  .502}{\textbf{9.7 }} & 8.3   & \textcolor[rgb]{ 0,  .502,  .502}{\textbf{9.4 }} & \textcolor[rgb]{ 0,  .502,  .502}{\textbf{10.6 }} & 8.8   & 10.4  & 9.9   & 1.8   & 7.1   & \textcolor[rgb]{ 1,  0,  0}{\textbf{10.4 }} & 2.3   & \textcolor[rgb]{ 0,  .502,  .502}{\textbf{7.3 }} & \textcolor[rgb]{ 0,  .502,  .502}{\textbf{9.9 }} & 5.3   & 8.4  \\
    GAGS \cite{peng2024gags}  & \textcolor[rgb]{ 0,  .502,  .502}{\textbf{14.6}}  & 8.7   & \textcolor[rgb]{ 0,  .502,  .502}{\textbf{13.6}}  & 4.8   & \textcolor[rgb]{ 0,  .502,  .502}{\textbf{10.3 }} & 5.8   & 8.4   & \textcolor[rgb]{ 0,  .502,  .502}{\textbf{16.3 }} & 12.7  & 8.6   & \textcolor[rgb]{ 0,  .502,  .502}{\textbf{12.8 }} & 11.1  & 7.0   & 6.0   & 6.8   & 8.7   & \textcolor[rgb]{ 0,  .502,  .502}{\textbf{10.8 }} & 10.0  \\
    Langsplat\cite{qin2024langsplat}  & 12.8  & \textcolor[rgb]{ 0,  .502,  .502}{\textbf{8.9}} & 11.9 & 9.0   & 8.6   & 9.2   & 8.9   & 16.2  & \textcolor[rgb]{ 0,  .502,  .502}{\textbf{13.0 }} & \textcolor[rgb]{ 0,  .502,  .502}{\textbf{11.6 }} & 11.6  & \textcolor[rgb]{ 0,  .502,  .502}{\textbf{11.6 }} & 6.4   & \textcolor[rgb]{ 0,  .502,  .502}{\textbf{6.4 }} & 6.4   & 9.7   & 10.3  & \textcolor[rgb]{ 0,  .502,  .502}{\textbf{10.4 }} \\
    Ours  & \textcolor[rgb]{ 1,  0,  0}{\textbf{15.0 }} & \textcolor[rgb]{ 1,  0,  0}{\textbf{12.1 }} & \textcolor[rgb]{ 1,  0,  0}{\textbf{14.8 }} & \textcolor[rgb]{ 1,  0,  0}{\textbf{13.8 }} & \textcolor[rgb]{ 1,  0,  0}{\textbf{18.6 }} & \textcolor[rgb]{ 1,  0,  0}{\textbf{16.9 }} & \textcolor[rgb]{ 1,  0,  0}{\textbf{11.0 }} & \textcolor[rgb]{ 1,  0,  0}{\textbf{16.4 }} & \textcolor[rgb]{ 1,  0,  0}{\textbf{14.6 }} & \textcolor[rgb]{ 1,  0,  0}{\textbf{13.0 }} & \textcolor[rgb]{ 1,  0,  0}{\textbf{13.1 }} & \textcolor[rgb]{ 1,  0,  0}{\textbf{14.0 }} & \textcolor[rgb]{ 0,  .502,  .502}{\textbf{9.2 }} & \textcolor[rgb]{ 1,  0,  0}{\textbf{11.0 }} & \textcolor[rgb]{ 1,  0,  0}{\textbf{10.4 }} & \textcolor[rgb]{ 1,  0,  0}{\textbf{12.4 }} & \textcolor[rgb]{ 1,  0,  0}{\textbf{14.2 }} & \textcolor[rgb]{ 1,  0,  0}{\textbf{14.1 }} \\
    \bottomrule
    \end{tabular}%
  \label{tab:adt_loc}%
\end{table*}%

\begin{table}[htbp]
  \centering
    \footnotesize
  \setlength{\tabcolsep}{2.5pt} % 设置列间距
  \caption{\textbf{Ablation Study on the ADT dataset.}}
    \begin{tabular}{ll|ccc|ccc}
    \toprule
    \multicolumn{2}{l|}{Component} & \multicolumn{6}{c}{Performance} \\
    \midrule
    Multi-view & \multicolumn{1}{l|}{Transinet } & \multicolumn{3}{c|}{Acc(\%)} & \multicolumn{3}{c}{IoU(\%)} \\
    Aggregation & \multicolumn{1}{l|}{Prediction} & \multicolumn{1}{l}{Static} & \multicolumn{1}{l}{Dynamic} & \multicolumn{1}{l|}{All} & \multicolumn{1}{l}{Static} & \multicolumn{1}{l}{Dynamic} & \multicolumn{1}{l}{All} \\
    \midrule
          &       & Static & Dynamic & All   & Static & Dynamic & All \\
          &       &26.9  &26.4  &27.7  &9.7   &10.3  &10.4  \\
    \checkmark &       & \textcolor[rgb]{ 0,  .502,  .502}{\textbf{31.3 }} & \textcolor[rgb]{ 0,  .502,  .502}{\textbf{31.2 }} & \textcolor[rgb]{ 0,  .502,  .502}{\textbf{33.3 }} & \textcolor[rgb]{ 0,  .502,  .502}{\textbf{11.9 }} & \textcolor[rgb]{ 0,  .502,  .502}{\textbf{14.1 }} & \textcolor[rgb]{ 0,  .502,  .502}{\textbf{13.6 }} \\
    \checkmark & \checkmark & \textcolor[rgb]{ 1,  0,  0}{\textbf{32.6 }} & \textcolor[rgb]{ 1,  0,  0}{\textbf{34.2 }} & \textcolor[rgb]{ 1,  0,  0}{\textbf{35.9 }} & \textcolor[rgb]{ 1,  0,  0}{\textbf{12.4 }} & \textcolor[rgb]{ 1,  0,  0}{\textbf{14.2 }} & \textcolor[rgb]{ 1,  0,  0}{\textbf{14.1 }} \\
    \bottomrule
    \end{tabular}%
  \label{tab:adt_abl}%
\end{table}%

\vspace{1mm}
\noindent \textbf{Baselines.} In addition to baseline models, we compare our method to three other state-of-the-art language embedded field methods: LERF~\cite{chen2024nerf}, LEGaussian~\cite{li2022language}, Semantics-Gaussian (SG)~\cite{guo2024semantic}, GAGS~\cite{peng2024gags} and Langsplat~\cite{qin2024langsplat}.

\subsection{Results on the ADT dataset}
We conducted open-vocabulary localization and 3D semantic segmentation experiments on the ADT dataset using an experimental setup similar to LangSplat~\cite{qin2024langsplat}. It is noteworthy that the ADT dataset includes off-the-shelf annotations for target motion types, categorizing targets as ``static" or ``dynamic".  Consequently, our evaluation metrics are computed separately for static targets, dynamic targets, and the combined set of all objects, while these motion labels are excluded during model training.

% \vspace{1mm}
\noindent \textbf{Quantitative Results.} The open-vocabulary location accuary results are reported in \cref{tab:adt_loc}. We observe that our method achieves an overall accuracy of 35.9\% in all the objects,  representing a 8.2\%  improvement over the second-best result obtained by LangSplat. On static targets, our algorithm achieves a 5.7\% improvement over the 2nd best results, whereas on dynamic targets, it demonstrates a 6.8\% increase in performance compared to GAGS.
Additionally, \cref{tab:adt_seg} reports the average Intersection over Union (IoU) results for 3D semantic segmentation, where our method achieves a 3.7\% improvement, underscoring the superiority of the proposed EgoSplat. Overall, the results indicate that our algorithm offers a significant advantage across a majority of scenarios for both static and dynamic targets.
Note that in the datasets, a single text query may correspond to multiple object masks, which can include both static and dynamic objects. Consequently, the evaluation value for all objects does not necessarily fall between those of static and dynamic objects.

% \vspace{1mm}
\noindent \textbf{Ablation Study.} We conduct ablations on the ADT dataset and report the average localization accuracy and  average IoU segmentation results in \cref{tab:adt_abl}. Without any of our proposed components, i.e., 3DGS with language features follow LangSplat ~\cite{qin2024langsplat}, performs the worst. Introducing multi-view aggregation improved localization accuracy by 5.6\% and semantic segmentation IoU by 3.2\% across all objects. The transient prediction module enhanced localization and segmentation accuracy for static targets by 1.3\% and 0.5\%, respectively. The overall accuracy and IoU increased by 2.6\% and 0.5\%, respectively.

\begin{figure*}[t]
  \centering
   \includegraphics[width=1\linewidth]{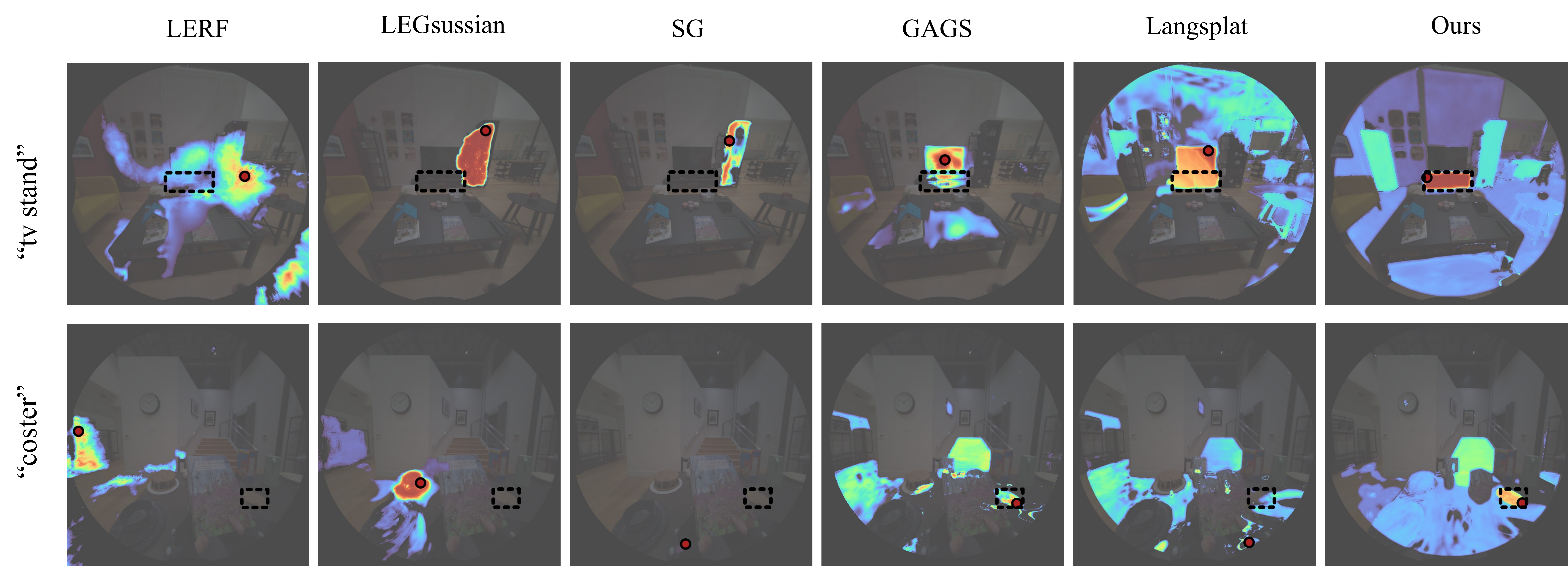}

   \caption{\textbf{Comparison of the visualization of open-vocabulary localization on the ADT dataset}. We selected ``tv stand" and ``coaster" for visualization. The red points are the model predictions and the black dashed bounding boxes denote the annotations.  It can be observed that our algorithm achieves the most accurate localization with the clearest boundaries.
   }
   \label{fig:adt_vis_loc}
\end{figure*}

\begin{figure*}[t]
  \centering
   \includegraphics[width=1\linewidth]{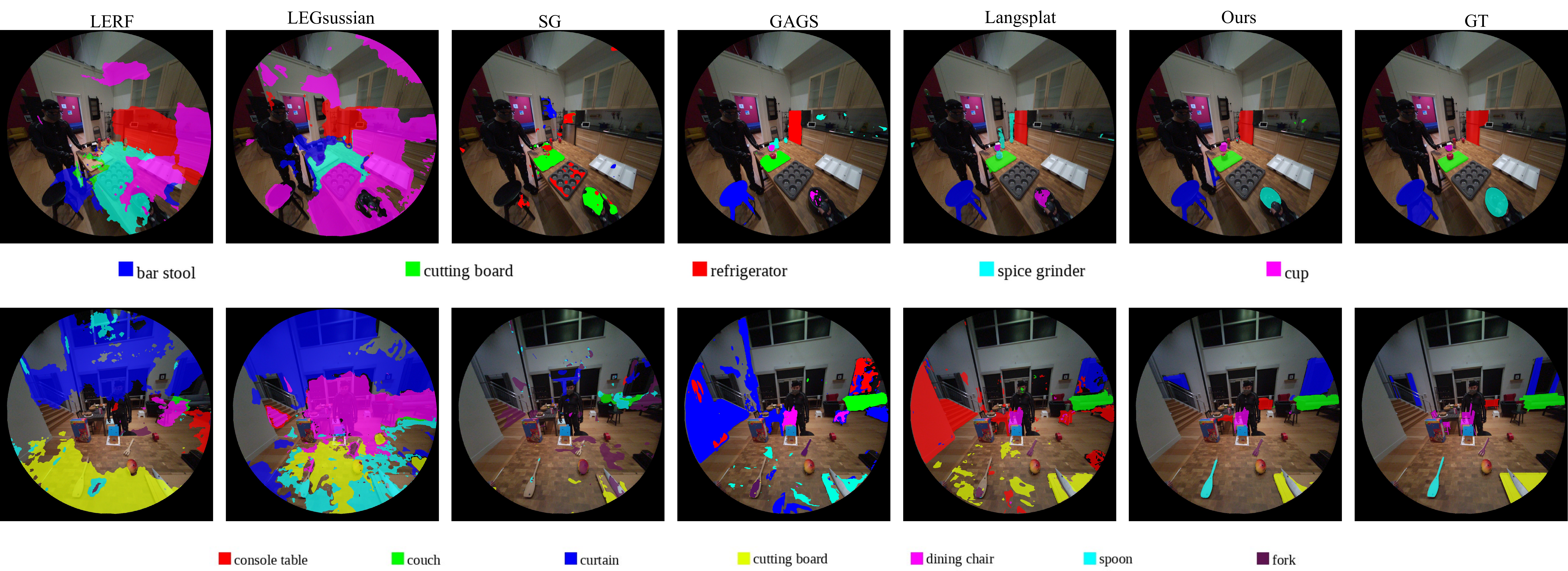}

   \caption{\textbf{Comparison of the visualization of open-vocabulary segmentation on the ADT dataset}. We selected several categories from two scenes for demonstration. It can be observed that our algorithm achieves the most accurate segmentation masks. 
   }
   \label{fig:adt_vis_seg}
   \vspace{-2mm}
\end{figure*}

% \vspace{1mm}
\noindent \textbf{Visualization Results.} Similar to the setup in LangSplat, we present the visualization results of object localization and semantic segmentation in \cref{fig:adt_vis_loc} and \cref{fig:adt_vis_seg}, respectively.
%Given that each scene in the ADT dataset comprises approximately 60 categories, we have selected a representative subset for visualization.
Our observations indicate that the activation regions produced by EgoSplat exhibit the most precise localization with the clearest boundaries. In terms of semantic segmentation, our approach also achieves the most complete and accurate delineation of boundaries.

\subsection{Results on the HOI4D Dataset}
% \vspace{1mm}
\noindent \textbf{Quantitative Results.} We conducted open-vocabulary localization and 3D semantic segmentation experiments on the HOI4D dataset, following a methodology similar to that used with the ADT dataset. The results are presented in \cref{tab:hoi4d}, which contains our EgoSplat method against baseline approaches. 
 It can be observed that EgoSplat showed a  a certain improvement in localization accuracy, and substantial superiority over other algorithms in 3D semantic segmentation. Notably, EgoSplat consistently achieved leading performance across all the evaluated algorithms. Additional experimental results on the HOI4D dataset are provided in the supplementary materials.

\begin{table}
  \centering
\footnotesize
  \caption{\textbf{Localization Accuracy and Segmentation IoU on the HOI4D dataset}.  The \textcolor[rgb]{ 1,  0,  0}{\textbf{1st}} and \textcolor[rgb]{ 0,  .502,  .502}{\textbf{2nd}} best results are highlighted.}
    \begin{tabular}{l|c|c}
    \toprule
          & ACC   & IoU \\
    \midrule
    LERF~\cite{li2022language}  & \textcolor[rgb]{ 0,  .502,  .502}{\textbf{67.5 }} & 21.9  \\
    LEGaussian~\cite{shi2024language} & 54.8  & 19.2  \\
    SG~\cite{guo2024semantic} & 51.0  & 33.5  \\
    GAGS~\cite{peng2024gags}& 55.0  & 40.5  \\
    Langsplat~\cite{qin2024langsplat} & 64.1  & \textcolor[rgb]{ 0,  .502,  .502}{\textbf{43.1 }} \\
    Ours  & \textcolor[rgb]{ 1,  0,  0}{\textbf{68.8 }} & \textcolor[rgb]{ 1,  0,  0}{\textbf{45.3 }} \\
    \bottomrule
    \end{tabular}%
  \label{tab:hoi4d}%
\end{table}%

\section{Conclusion}
\label{sec:con}
We introduced EgoSplat for open-vocabulary egocentric scene understanding using language-embedded 3DGS. To address the challenges, like frequent occlusions and dynamic artifacts disrupting multi-view consistency in egocentric perspectives, we developed a multi-view instance semantics extraction mechanism that aggregates semantic features across high-quality views with the assistance of  advanced segmentation and tracking method. Additionally, we proposed an instance-aware spatial-temporal transient prediction module to predict the motion elements completely  and continuously  and reduce dynamic artifacts effectively. Evaluated on two prominent egocentric datasets, our framework outperforms existing state-of-the-art methods in localization and segmentation tasks, setting a new benchmark for dynamic scene understanding in egocentric videos.  Please see the supplementary details for limitations and future work.

%-------------------------------------------------------------------------

{
    \small
    \bibliographystyle{ieeenat_fullname}
    \bibliography{main}
}


% \documentclass[10pt,twocolumn,letterpaper]{article}
% \usepackage[final]{iccv}

% % Include other packages here, before hyperref.
% \usepackage{graphicx}
% \usepackage{amsmath}
% \usepackage{amssymb}
% \usepackage{booktabs}

% \usepackage{multirow}
% \usepackage{makecell}

% \usepackage{colortbl}
% \usepackage{graphicx}
% \usepackage{caption}
% \usepackage{float}
% Import additional packages in the preamble file, before hyperref
% \input{preamble}

% If you comment hyperref and then uncomment it, you should delete
% egpaper.aux before re-running latex.  (Or just hit 'q' on the first latex
% run, let it finish, and you should be clear).
% \definecolor{cvprblue}{rgb}{0.21,0.49,0.74}
% \usepackage[pagebackref,breaklinks,colorlinks,allcolors=cvprblue]{hyperref}

% If you wish to avoid re-using figure, table, and equation numbers from
% the main paper, please uncomment the following and change the numbers
% appropriately.
%\setcounter{figure}{2}
%\setcounter{table}{1}
%\setcounter{equation}{2}

% If you wish to avoid re-using reference numbers from the main paper,
% please uncomment the following and change the counter value to the
% number of references you have in the main paper (here, 100).
%\makeatletter
%\apptocmd{\thebibliography}{\global\c@NAT@ctr 100\relax}{}{}
%\makeatother

\pagestyle{plain} % 确保显示页码

\renewcommand{\thetable}{S.\arabic{table}}
\renewcommand{\thefigure}{S.\arabic{figure}}

% \begin{document}

% \title{EgoSplat: Open-Vocabulary Egocentric Scene Understanding with Language Embedded 3D Gaussian Splatting}

\twocolumn[{%
  \renewcommand\twocolumn[1][]{#1}%
  \maketitlesupplementary
\begin{center}
    \captionsetup{type=figure}
    \includegraphics[width=.8\textwidth]{sec/fig/sup_view_diff.pdf}
    % \captionof{figure}{Test caption}
   \caption{\textbf{Visualization of the Occlusion Case. } (b) CLIP fails to accurately recognize the ``chair'' when it is occluded by a moving actor. (c) Inaccurate features are filtered out through multi-view feature aggregation, and view-independent features are aggregated, resulting in (d) correct semantics.}
   \label{fig:sup_view_diff}
\end{center}
}]
% \maketitlesupplementary

\thispagestyle{empty}
\appendix

% \tableofcontents

%%%%%%%%% BODY TEXT - ENTER YOUR RESPONSE BELOW

% % \section{Additional Results}
% \label{sec:sup_data}

\section{Performance in Challenging Cases}

% \noindent \textbf{Occlusion.}
\subsection{Occlusion Case}
\cref{fig:sup_view_diff} illustrates a scenario where occlusion leads to semantic recognition errors. Specifically, when relying solely on view-specific semantic information (as depicted in the lower part of (b)), CLIP fails to accurately recognize the ``chair'' when it is occluded by a moving actor. By implementing our proposed multi-view consistent instance feature aggregation, view-specific features are filtered out while view-independent features are effectively aggregated, resulting in correct semantic recognition.  Our algorithm successfully mitigates the challenges posed by frequent occlusions, ensuring robust performance in complex scenarios.

\subsection{Transient Case}

\begin{figure}[t]
  \centering
   \includegraphics[width=1\linewidth]{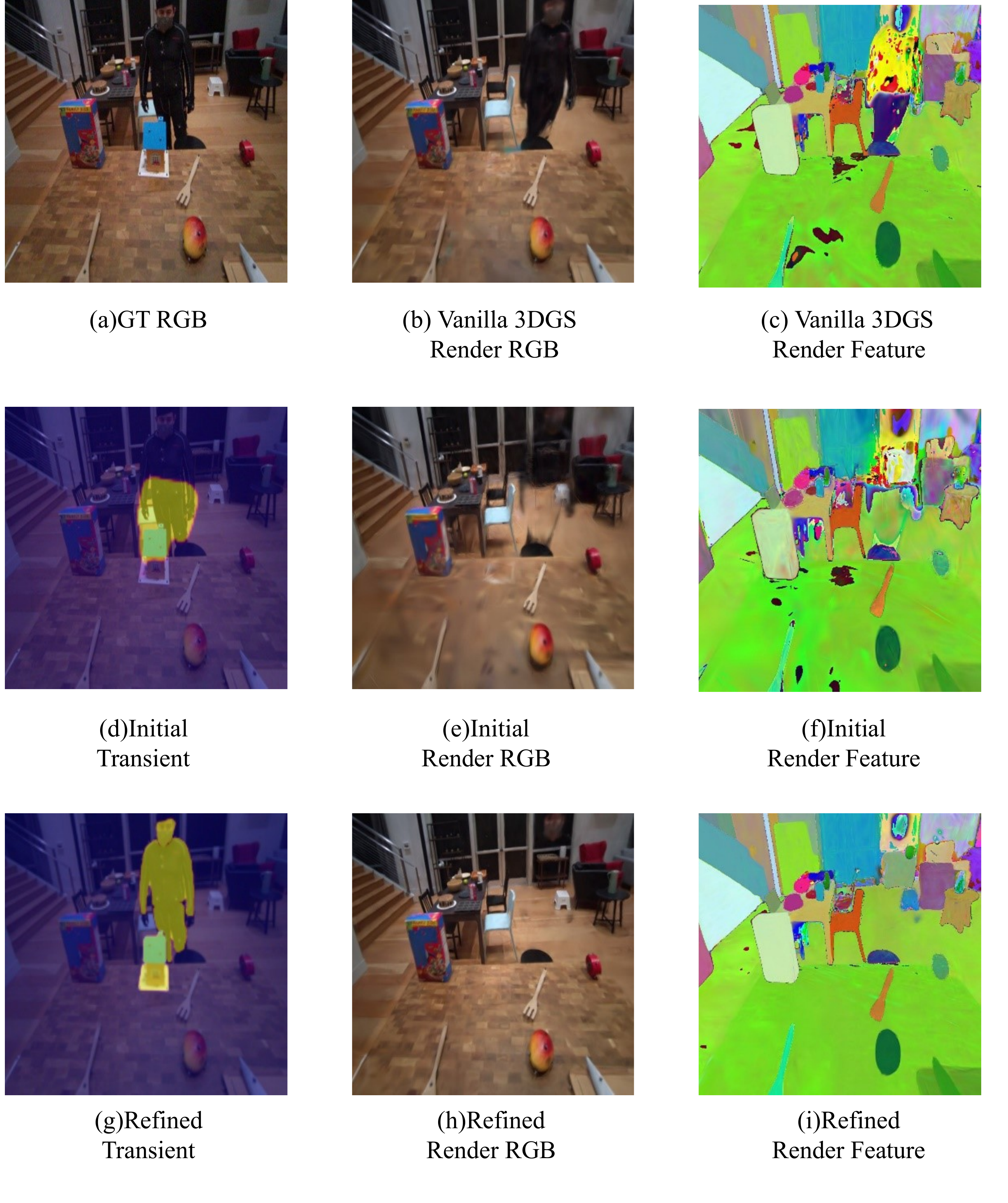}

   \caption{\textbf{Visualization of the Transient Case. } (a) Ground truth RGB image.
(b-c) Vanilla 3DGS rendering without transient processing, showing artifacts.
(d-f) Initial stage results with incomplete transient prediction but reduced artifacts.
(g-i) Refinement stage results with complete transient prediction and artifacts almost entirely eliminated. } 
   
   \label{fig:sup_trans_case}
\end{figure}

\cref{fig:sup_trans_case} demonstrates a typical instance of dynamic actor-object interactions in egocentric videos. Without transient processing, as shown in the first row, the rendered RGB images and language features produced by vanilla 3DGS exhibit significant artifacts and a lack of spatial consistency.
The second row presents results of the initial transient prediction stage, where significant artifact reduction is observed despite incomplete transient predictions. Finally, the refinement stage (third row) achieves near-complete transient predictions, effectively eliminating artifacts and producing spatially consistent rendered RGB images and language features. This progression underscores the critical role of refinement in improving the quality of transient predictions and enhancing the fidelity of rendered outputs.

\section{Ablation Studies on Transient Prediction}
% \noindent \textbf{Ablation Studies on Transient Prediction.}\\
As illustrated in \cref{tab:sup_ab_trans}, we conducted experiments on scene {\footnotesize
\texttt{Apartment\_release\_multiskeleton\_party\_seq121\_M1292}} of the ADT dataset. The results indicate that  performing transient prediction solely during the initial stage yielded limited improvements in localization performance; indeed, the segmentation IoU marginally decreased. This phenomenon is likely due to the increased complexity of optimization. However, the introduction of the refinement stage led to notable improvements in both localization and segmentation metrics, particularly under static conditions.
\begin{table}[htbp]
  % \centering
  \centering
  \footnotesize
  \setlength{\tabcolsep}{2.5pt} % 设置列间距
  \caption{Ablation Study of the Transient Prediction.}
    \begin{tabular}{ll|ccc|ccc}
    \toprule
    \multicolumn{2}{l|}{Component} & \multicolumn{6}{c}{Performance} \\
    \midrule
    \multicolumn{1}{c}{\multirow{2}[1]{*}{initial-stage}} & \multicolumn{1}{c|}{\multirow{2}[1]{*}{refine-stage}} & \multicolumn{3}{c|}{Acc(\%)} & \multicolumn{3}{c}{mIoU(\%)} \\
          &       & Static & Dynamic & All   & Static & Dynamic & All \\
    \midrule
          &       & 32.4  & \textcolor[rgb]{ 0,  .502,  .502}{\textbf{29.4}} & 34.6  & \textcolor[rgb]{ 0,  .502,  .502}{\textbf{13.5}} & \textcolor[rgb]{ 1,  0,  0}{\textbf{12.3}} & \textcolor[rgb]{ 0,  .502,  .502}{\textbf{14.0}} \\
    \checkmark &       & \textcolor[rgb]{ 0,  .502,  .502}{\textbf{32.7}} & \textcolor[rgb]{ 1,  0,  0}{\textbf{29.4}} & \textcolor[rgb]{ 0,  .502,  .502}{\textbf{34.8}} & 11.9  & 11.5  & 12.6  \\
    \checkmark & \checkmark & \textcolor[rgb]{ 1,  0,  0}{\textbf{34.2}} & 28.3  & \textcolor[rgb]{ 1,  0,  0}{\textbf{35.2}} & \textcolor[rgb]{ 1,  0,  0}{\textbf{15.0}} & \textcolor[rgb]{ 0,  .502,  .502}{\textbf{12.1}} & \textcolor[rgb]{ 1,  0,  0}{\textbf{14.8}} \\
    \bottomrule
    \end{tabular}%
  \label{tab:sup_ab_trans}%
\end{table}%

\section{More Visualization Results}
\begin{figure*}[t]
  \centering
   \includegraphics[width=0.6\linewidth]{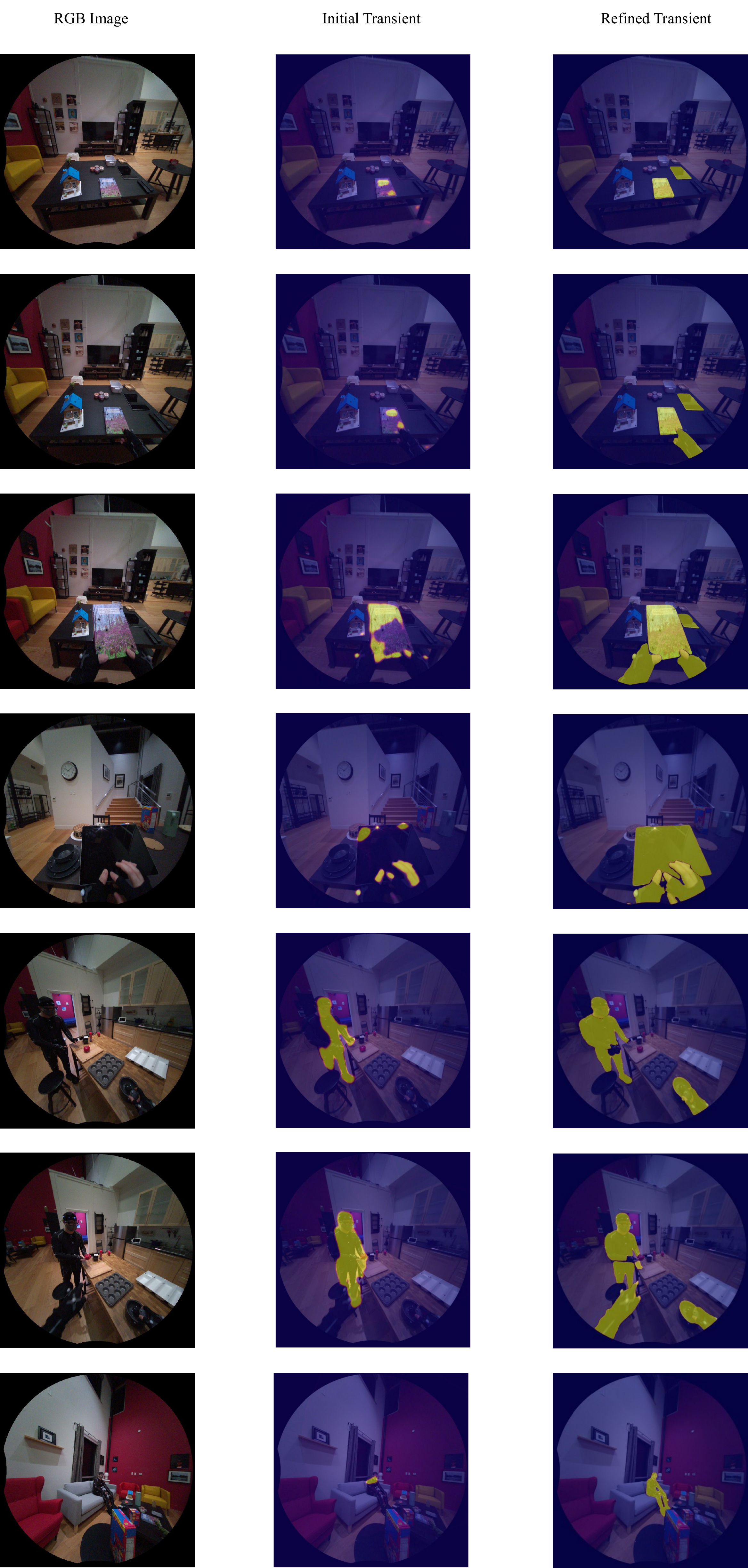}

   \caption{\textbf{Visualization of the Transient Prediction}. The yellow regions indicate the parts predicted as transient.}
   \label{fig:sup_trans}
\end{figure*}

\vspace{1mm} 
\noindent \textbf{Visualization of the Transient Prediction.} The visualization of transient predictions on the ADT dataset is shown in \cref{fig:sup_trans}, where yellow regions represent areas identified as transient. Initially, the transient mask appears spatially incomplete and is often absent in certain frames over time. Following the refinement stage, the transient prediction becomes both spatially complete and temporally consistent, demonstrating the effectiveness of the proposed refinement process.

\begin{figure*}[t]
  \centering
   \includegraphics[width=1\linewidth]{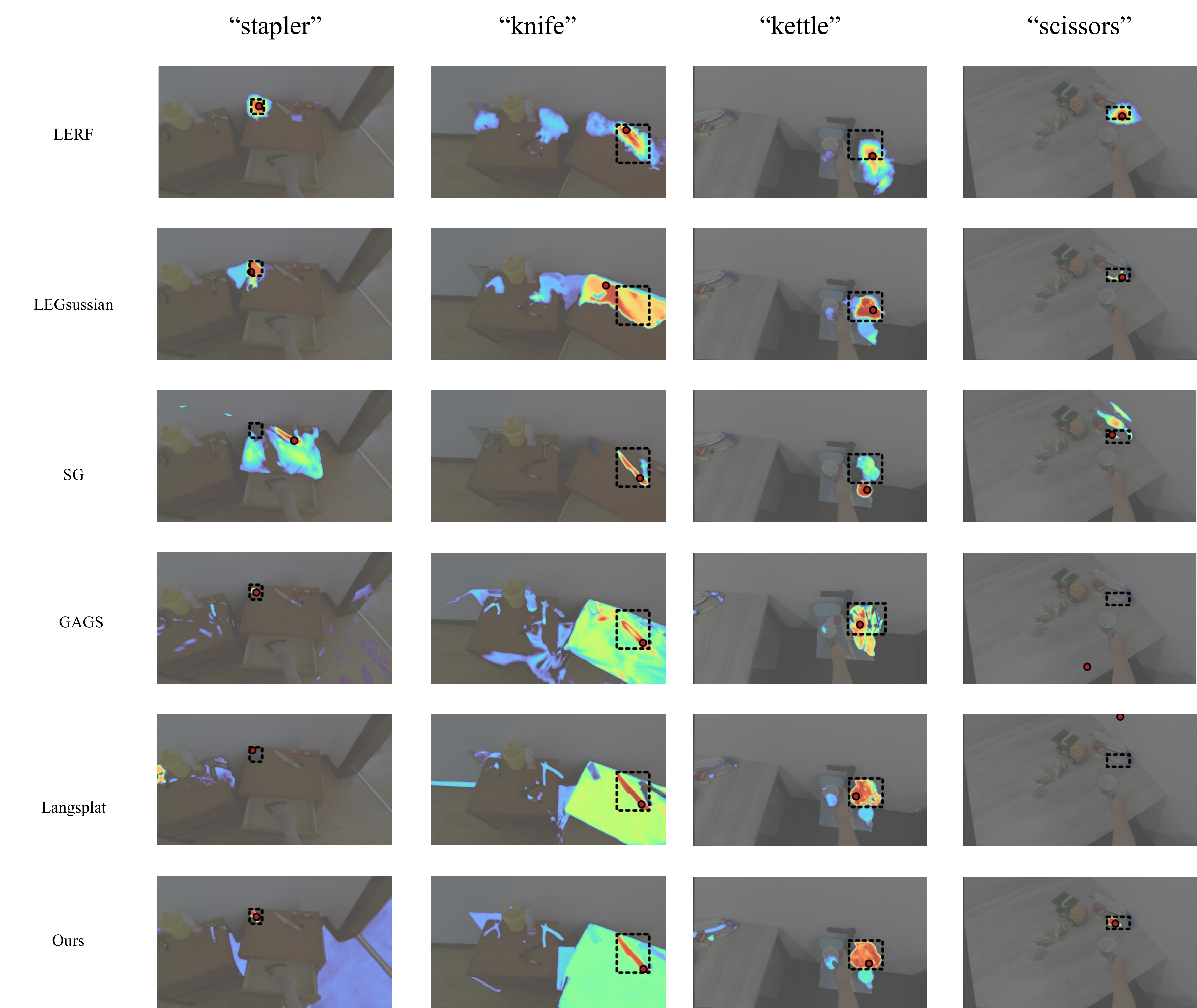}

   \caption{\textbf{Comparison of the visualization of open-vocabulary localization on the HOI4D dataset}. We selected ``stapler", ``knife", "kettle" and "scissors" for visualization. The red points are the model predictions and the black dashed bounding boxes denote the annotations.  It can be observed that our algorithm achieves the most accurate localization with the clearest boundaries.
   }
   \label{fig:sup_hoi4d_loc}
\end{figure*}

\vspace{1mm} 
\noindent \textbf{Visualization of Localization on  HOI4D.} Visualization results for open-vocabulary localization on the HOI4D dataset are shown in \cref{fig:sup_hoi4d_loc}. Our proposed EgoSplat demonstrates superior localization accuracy compared to other methods, effectively capturing activation regions that are highly aligned with the target shapes.

\begin{figure*}[t]
  \centering
   \includegraphics[width=0.6\linewidth]{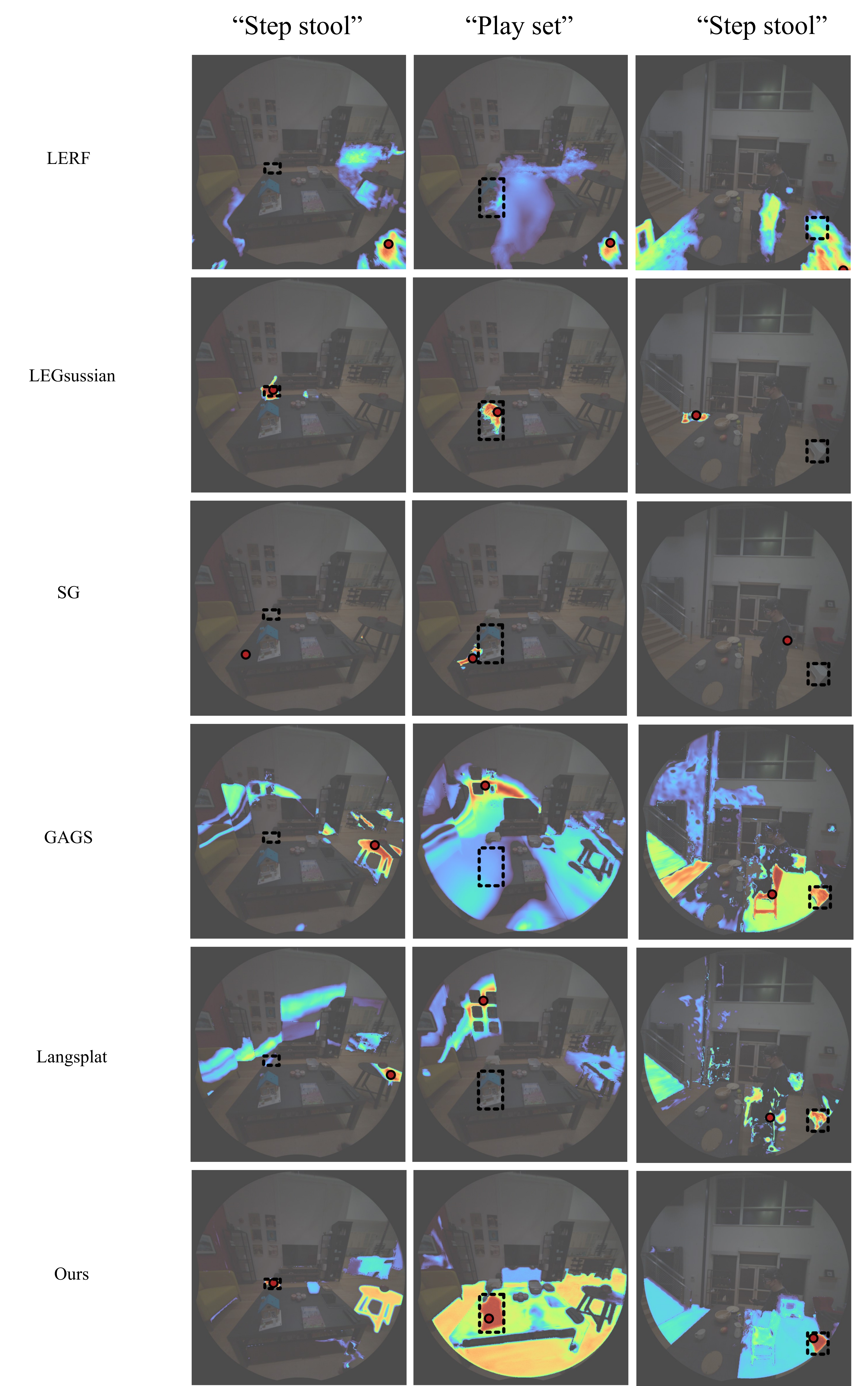}

   \caption{\textbf{More comparison of the visualization of open-vocabulary localization on the ADT dataset}. 
   }
   \label{fig:sup_adt_loc}
\end{figure*}

\vspace{1mm} 
\noindent \textbf{Extended Qualitative Comparison of Localization on ADT.} Further comparisons of the visualization of open-vocabulary localization on the ADT dataset are shown in \cref{fig:sup_adt_loc}. The EgoSplat method achieves the most accurate localization results, particularly with small objects such as ``step stool''.

\begin{figure*}[p]
    \centering
    \includegraphics[width=\textwidth]{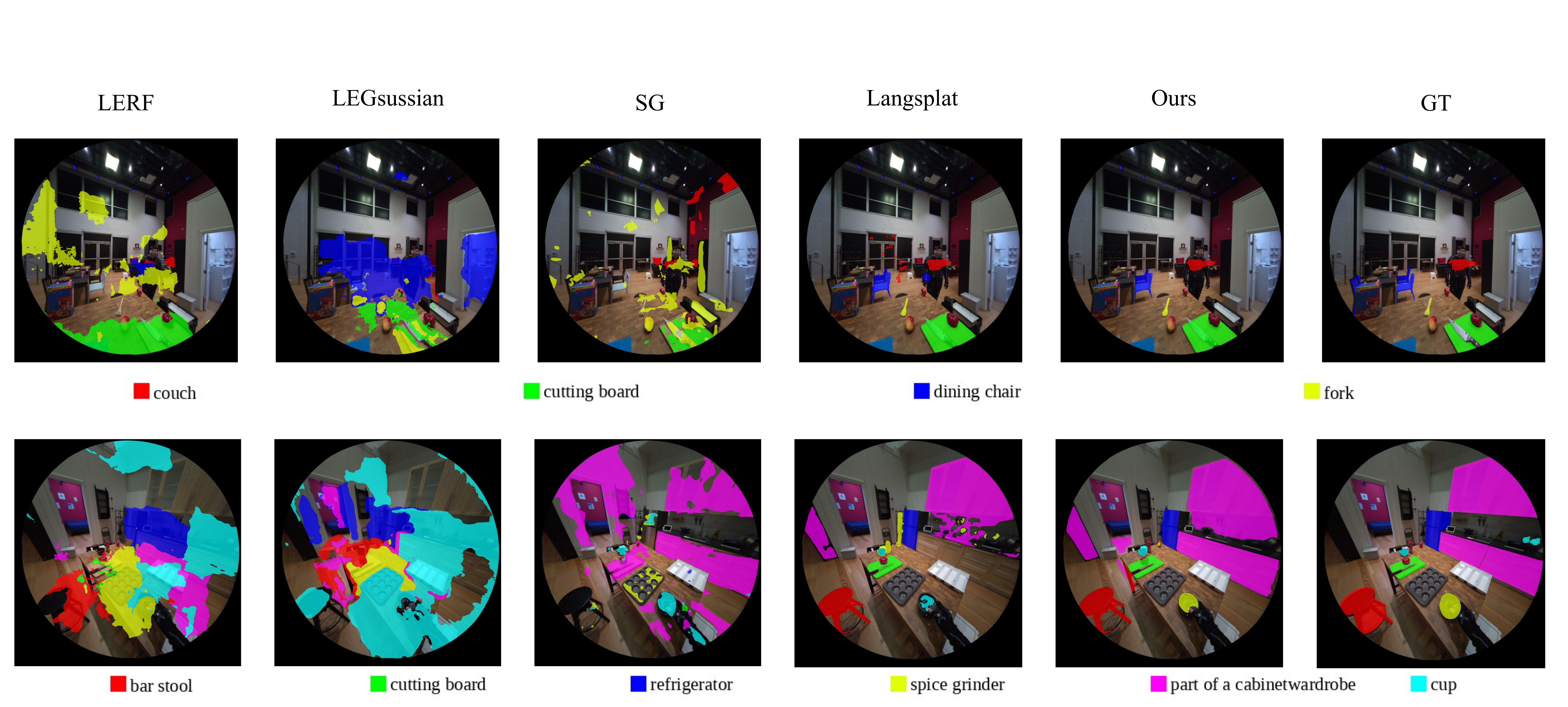}
    \caption{\textbf{More comparison of the visualization of open-vocabulary segmentation on the ADT dataset}.}
    \label{fig:sup_adt_seg}
\end{figure*}

\vspace{1mm} 
\noindent \textbf{Extended Qualitative Comparison of Segmentation on ADT.} Further visual comparisons of open-vocabulary segmentation on the ADT dataset are shown in \cref{fig:sup_adt_seg}. EgoSplat demonstrates superior segmentation accuracy, producing precise masks across targets of varying sizes.

\section{Additional Implementation Details}
\label{sec:sup_im}
\subsection{Details of the Proposed EgoSplat}
\vspace{1mm}
\noindent \textbf{Data Preprocessing.} For the ADT dataset, we employed the off-the-shelf point cloud and camera parameters. For the HOI4D dataset, we generated the point cloud and camera parameters utilizing COLMAP~\cite{schoenberger2016sfm}, a widely used Structure-from-Motion (SfM) tool.

\vspace{1mm}
\noindent \textbf{Feature Aggregation.}
For each text query, we generate three relevancy maps using our trained 3D language Gaussians, each corresponding to a distinct semantic level defined by SAM,  follow the selection of ~\cite{qin2024langsplat}. The strategies for selecting the most appropriate semantic level and obtaining predictions for different tasks are also consistent with those employed in LangSplat~\cite{qin2024langsplat}. The weight parameter $\lambda_1$ for the semantic consistency loss $\mathcal{L}_{\text{F}}$ is set to $1$. 

\vspace{1mm}
\noindent \textbf{Transient Prediction.} For transient prediction, during the initial stage, the distance functions \( d_I \) and \( d_F \) used for RGB  and feature in 3DGS are set to the L2 distance. During the refinement stage, these distance functions are adjusted to the L1 distance.The threshold \( \mathcal{T} \) is set to $0.02$ for the ADT dataset, whereas for the HOI4D dataset, this value is adjusted to $0.1$. The weight parameter $\lambda_2$ for the regularization term is set to $0.2$. 

\vspace{1mm}
\noindent \textbf{Optimization.} The 3DGS  models are trained using the Adam optimizer, following the same configuration and density control schedule as outlined in ~\cite{gu2025egolifter}. The transient prediction network is optimized separately with the Adam optimizer, employing an initial learning rate of $1 \times 10^{-5}$.

\subsection{Details of Comparisons}
\vspace{1mm}
\noindent \textbf{LERF~\cite{kerr2023lerf}.} LERF is implemented using Nerfstudio~\cite{nerfstudio}, and the model architecture aligns with those in the official codebase. Following the default settings,  we trained LERF for 30,000 steps using the Adam optimizer. An exponential learning rate scheduler was employed, decreasing the learning rate from \(10^{-2} \) to  \(10^{-3} \) over the initial 5,000 training steps.  

\vspace{1mm}
\noindent \textbf{LEGaussian~\cite{shi2024language}.}  The model architecture is consistent with those in the official codebase. For all scenes, the size \(N\) of the discrete language feature is set to 128. 

\vspace{1mm}
\noindent \textbf{SG~\cite{guo2024semantic}.} For the per-pixel semantic maps, we selected the OpenSeg~\cite{ghiasi2022scaling} features, which demonstrated the best performance on the ADT dataset. The section of the paper on dynamic Gaussians with temporal information has not been verified due to the absence of open-source code. The remaining model architecture and training settings are consistent with those in the official codebase.

\vspace{1mm}
\noindent \textbf{GAGS~\cite{peng2024gags}. }The 512-dimensional CLIP features were compressed to 3 dimensions to conserve GPU memory, with both RGB fields and feature training undergoing 30,000 iterations each. The remaining model architecture and training settings are consistent with those in the official codebase.

\vspace{1mm}
\noindent \textbf{Langsplat~\cite{qin2024langsplat}.} The model architecture and training settings conform to the official codebase of the LERF dataset. Both RGB and language features are trained sequentially for 30,000 iterations each. An autoencoder implemented using MLP,  compresses the 512-dimensional CLIP features into 3-dimensional latent representations to reduce memory consumption.

\subsection{Benchmark Details}
\label{sec:sup_data}
% \subsection{ADT dataset}
\vspace{1mm}
\noindent \textbf{ADT~\cite{sun2023aria}.}

Egolifter~\cite{gu2025egolifter} selected 13 scenes from the ADT dataset for interactive segmentation evaluation, excluding sequences with excessively narrow baselines unsuitable for 3D reconstruction and those lacking segmentation annotations for human bodies.
From these, we randomly selected 5 scenes to serve as the evaluation set for open-vocabulary segmentation and localization. 
The evaluation set of each scene in ADT was constructed through equidistant sampling, ensuring that the distribution of viewpoint variations is consistent with that of the entire dataset.
% , as shown in \cref{fig:reb_1}.
The selected sequences, referred to as scenes 1 through 5 in this paper, are listed below: \\
{\footnotesize
\texttt{Apartment\_release\_multiskeleton\_party\_seq121\_M1292} \\
\texttt{Apartment\_release\_multiskeleton\_party\_seq125\_M1292} \\
\texttt{Apartment\_release\_multiskeleton\_party\_seq123\_M1292} \\
\texttt{Apartment\_release\_multiuser\_party\_seq140\_M1292} \\
\texttt{Apartment\_release\_work\_skeleton\_seq140\_M1292}} \\

The image resolution for the ADT dataset is $1408\times 1408$, and all algorithms were trained at this resolution.

\vspace{1mm} 
\noindent \textbf{HOI4D~\cite{Liu_2022_CVPR}.}
We randomly selected 5 sequences from HOI4D, each featuring interactions between humans and the environment. The names of the 5 selected sequences are listed as follows: \\
{\footnotesize
\texttt{ZY20210800001\_H1\_C11\_N58\_S376\_s02\_T2} \\
\texttt{ZY20210800001\_H1\_C11\_N58\_S376\_s05\_T2} \\
\texttt{ZY20210800001\_H1\_C7\_N29\_S272\_s03\_T2} \\
\texttt{ZY20210800003\_H3\_C9\_N41\_S309\_s02\_T2} \\
\texttt{ZY20210800004\_H4\_C2\_N44\_S364\_s05\_T6}}\\
The off-the-shelf 2D panoptic segmentation annotations in HOI4D were deemed insufficiently detailed for open-vocabulary localization and segmentation tasks. To address this, we re-annotated the selected 5 scenes from the dataset, creating an evaluation benchmark designed for these tasks. We provided 2D instance mask annotations for 37 object categories across five scenes. This benchmark will be publicly released.
The image resolution for the HOI4D dataset is $1889\times 1056$, and all algorithms were trained at this resolution.

\section{Limitations and Future Work}
Our EgoSplat model relies on promptable visual foundation models for segmentation and tracking. Although it achieves excellent performance in semantic consistency and transient processing, one limitation is that incorrect tracking results can lead to degradation of target semantics. 
Exploring the potential of EgoSplat under challenging segmentation and tracking conditions remains desired, which is left to future work.
Additionally, EgoSplat requires a two-stage transient prediction process. Relying solely on the initial stage may slightly decrease open-vocabulary segmentation performance (\cref{tab:sup_ab_trans}), possibly due to the difficulty of jointly optimizing the transient prediction network and 3DGS. This two-stage training also slightly increases the overall training time. Integrating the initial and refinement stages could enhance both the performance and efficiency of the model, which is an avenue worth exploring. Finally, EgoSplat has primarily been evaluated in indoor environments, extending evaluations to outdoor environments and cross-domain scenarios could ensure broader applicability.

% %%%%%%%%% REFERENCES
% {
%     \small
%     \bibliographystyle{ieeenat_fullname}
%     \bibliography{main}
% }

\clearpage
\clearpage

% \end{document}

\end{document}